\documentclass[conference]{IEEEtran}
\IEEEoverridecommandlockouts
\usepackage{cite}
\usepackage{amsmath,amssymb,amsfonts}
\usepackage{algorithmic}
\usepackage{graphicx}
\usepackage{textcomp}
\usepackage{xcolor}
\usepackage{enumitem}
\usepackage[a4paper, total={184mm,239mm}]{geometry}
\def\BibTeX{{\rm B\kern-.05em{\sc i\kern-.025em b}\kern-.08em
    T\kern-.1667em\lower.7ex\hbox{E}\kern-.125emX}}
\begin{document}

\title{Navigating the Unknown: Uncertainty-Aware Compute-in-Memory Autonomy of Edge Robotics
\thanks{This work was supported in part by COGNISENSE, one of seven centers in JUMP 2.0, a Semiconductor Research Corporation (SRC) program sponsored by DARPA, and NSF CAREER Award \#2046435.}
}

\author{Nastaran Darabi, Priyesh Shukla, Dinithi Jayasuriya, Divake Kumar, Alex C. Stutts, and Amit Ranjan Trivedi \\
AEON Lab, University of Illinois Chicago (UIC), Chicago, IL, Email: amitrt@uic.edu
}

\maketitle

\begin{abstract}
This paper addresses the challenging problem of energy-efficient and uncertainty-aware pose estimation in insect-scale drones, which is crucial for tasks such as surveillance in constricted spaces and for enabling non-intrusive spatial intelligence in smart homes. Since tiny drones operate in highly dynamic environments, where factors like lighting and human movement impact their predictive accuracy, it is crucial to deploy uncertainty-aware prediction algorithms that can account for environmental variations and express not only the prediction but also \textit{confidence in the prediction}. We address both of these challenges with Compute-in-Memory (CIM) which has become a pivotal technology for deep learning acceleration at the edge. While traditional CIM techniques are promising for energy-efficient deep learning, to bring in the robustness of uncertainty-aware predictions at the edge, we introduce a suite of novel techniques: \textit{First}, we discuss CIM-based acceleration of Bayesian filtering methods uniquely by leveraging the Gaussian-like switching current of CMOS inverters along with co-design of kernel functions to operate with extreme parallelism and with extreme energy efficiency. \textit{Secondly}, we discuss the CIM-based acceleration of variational inference of deep learning models through probabilistic processing while unfolding iterative computations of the method with a compute reuse strategy to significantly minimize the workload. \textit{Overall}, our co-design methodologies demonstrate the potential of CIM to improve the processing efficiency of uncertainty-aware algorithms by orders of magnitude, thereby enabling edge robotics to access the robustness of sophisticated prediction frameworks within their extremely stringent area/power resources.
\end{abstract}

\begin{IEEEkeywords}
Compute-in-memory, Edge robotics, Autonomous Navigation, Visual Odometry, Particle Filtering
\end{IEEEkeywords}

\section{Introduction}
At the core of the self-navigation of autonomous drones and robots lies the fundamental task of continuously ascertaining their position and orientation, often referred to as their pose estimation \cite{kendall2015posenet,darabi2023starnet}. This task is indispensable for various autonomy tasks such as path planning and obstacle avoidance. Furthermore, pose estimation must remain reliable and resilient even under dynamic changes to the vehicle's operational environment. For instance, in indoor settings, the environment is subjected to ongoing alterations, including the movement of people and fluctuations in lighting conditions which significantly impact the accuracy and effectiveness of drone localization \cite{skrzypczynski2017mobile}. Moreover, in Fig. 1, computing uncertainties also arise due to non-idealities (such as process variability) and noise in the underlying hardware. For robotics applications where perception-action loops are tightly coupled, these various sources of uncertainties can intricately interact and can dramatically degrade the quality of predictions. Therefore, techniques that not only provide real-time pose estimation but also account for various uncertainties in their predictions are imperative for risk-aware decision-making \cite{shukla2023robust, parente2023conformalized}.

Simultaneously, there is an ongoing imperative to reduce the computational and storage requirements of autonomous operations while maintaining predictive robustness in emerging applications, particularly for \textit{autonomous insect-scale drones}. Remarkable progress has been achieved in the field of drone miniaturization, as evidenced by various studies \cite{hsiao2023heading, afakh2023study}, resulting in drones that weigh less than a gram and are smaller than a penny! However, as these drones shrink in size, their payload capacity is inherently limited, often constrained to accommodate only a tiny battery. Consequently, it becomes crucial to minimize the power consumption associated with on-board processing tasks. Although leveraging cloud resources for offloading on-board processing is an option, many environments present challenges with limited or no connectivity. Furthermore, the continuous transmission of high-resolution camera data to the cloud can also incur significant costs.

\begin{figure}[t!]
    \centering
    \includegraphics[width=\linewidth]{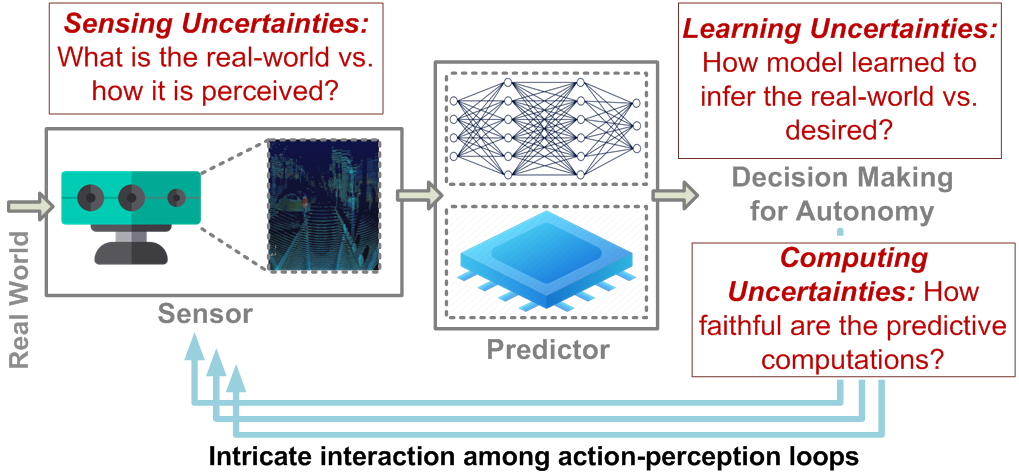}
    \caption{
Various sources of uncertainties arise in typical autonomous systems such as the inability to perceive the application domain reliably, discrepancies in the trained and applied environment, and sources of non-idealities such as noise and process variability impacting the reliability of processing.}
    \label{fig:enter-label}
\end{figure}

Compute-in-memory (CIM) has emerged as a promising technology for advancing ultra-low-power deep learning by integrating memory and processing in the same module, thereby significantly reducing the energy and time required for data movement between separate memory and processor units, which is a major bottleneck for deep learning computations in conventional computing architectures. Moreover, CIM-based deep learning can also minimize necessary processing steps by \textit{leveraging physics for computing}. For example, based on charge or current-domain representation of operands, additions of weight-input product terms can be realized simply on a wire by leveraging Kirchoff's law \cite{nasrin2023memory, yu2021compute}.  While CIM has revolutionized deterministic deep learning, its application in probabilistic and Bayesian inference remains largely untapped. As edge intelligence increasingly proliferates to scenarios demanding risk-averse, real-time predictions--such as in autonomous vehicles--the need for not just resource efficiency but also heightened robustness becomes paramount. 

In this article, we explore the significant potential of extending CIM to Bayesian and probabilistic inference, applicable to both deep learning and conventional signal processing algorithms where various sources of predictive uncertainties can be systematically accounted for. At the heart of our various proposals is a co-design methodology, which synergistically optimizes both the underlying hardware and computational models. This approach is crucial in maximizing the efficacy of the overall computing substrate, thereby unlocking new possibilities for intelligent, robust computing.

Our work makes the following key contributions:
\begin{itemize}[noitemsep,topsep=0pt,partopsep=0pt,leftmargin=*]
    \item We discuss CIM processing of particle filtering for drone localization, which employs a probabilistic approach to accurately determine a drone's position and orientation by continuously updating a set of hypotheses about its state. While the workload of particle filtering can be significant for complex and extensive environments, we demonstrate how the Gaussian-like switching current of CMOS inverters, along with a co-design of map models on CIM substrate can be leveraged to remarkably minimize the resources required for large-scale particle filtering. 
    \item We also discuss the CIM-based acceleration of Bayesian Deep Learning (BDL). In particular, exploring a CIM-based implementation of a variational inference procedure for BDL, called MC-Dropout, we demonstrate the unique opportunities for CIM to considerably minimize the workload of the Monte Carlo-based inference procedure. The discussed CIM modules not only store and operate on weight matrices, but also emulate sampling-based iterations of BDL by leveraging noise sources. Furthermore, our CIM-based BDL processing leverages opportunities such as compute reuse and optimal sample ordering, to significantly minimize the resource demands of BDL—bringing its requirements close to the energy/resource budget of edge devices.    
\end{itemize}

In essence, the paper compiles a range of insights derived from our previous research \cite{shukla2021ultralow, shukla2022mc, shukla2023robust, stutts2023lightweight, darabi2023adc} and underscores the significance of co-designing CIM with Bayesian algorithms as a versatile framework capable of substantially augmenting their implementation efficiency. Sec. II discusses CIM-based particle filtering for ultra-low-power drone localization by employing an iterative, reasoning-based prediction procedure. Sec. III presents CIM-based acceleration of MC-Dropout and characterization on visual odometry (VO).  Sec. IV concludes.  

\begin{figure*}[t]
    \centering
    \includegraphics[width=\linewidth]{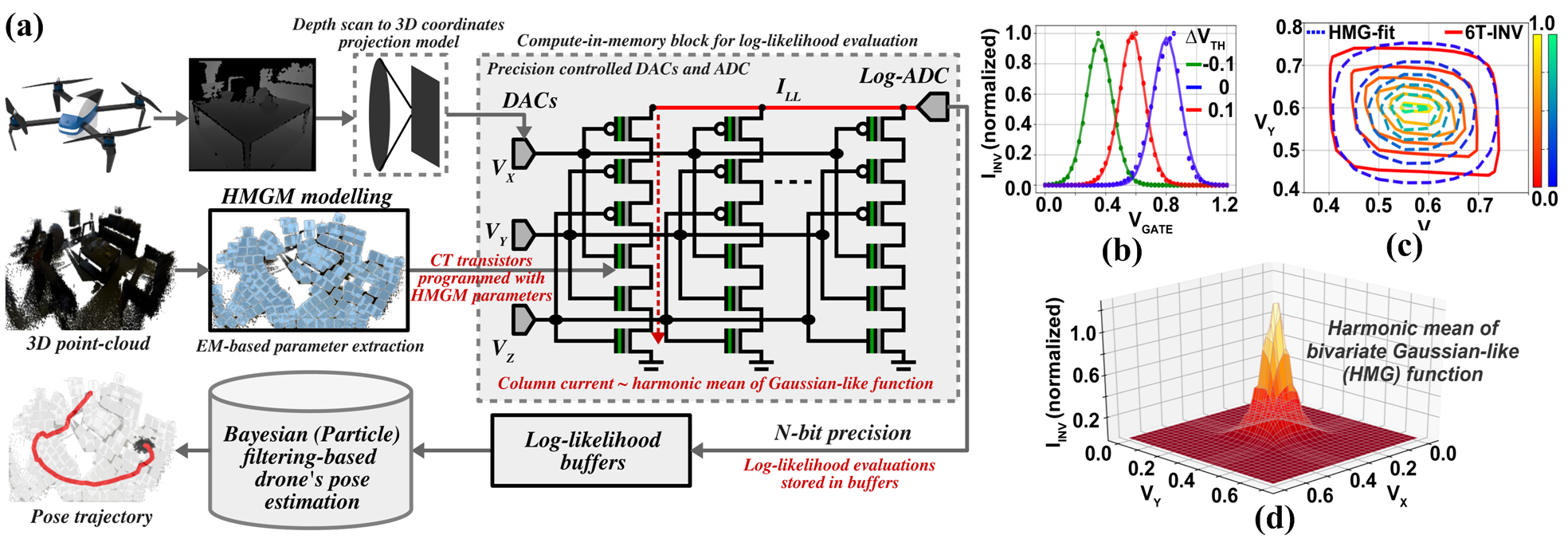}
    \includegraphics[width=\linewidth]{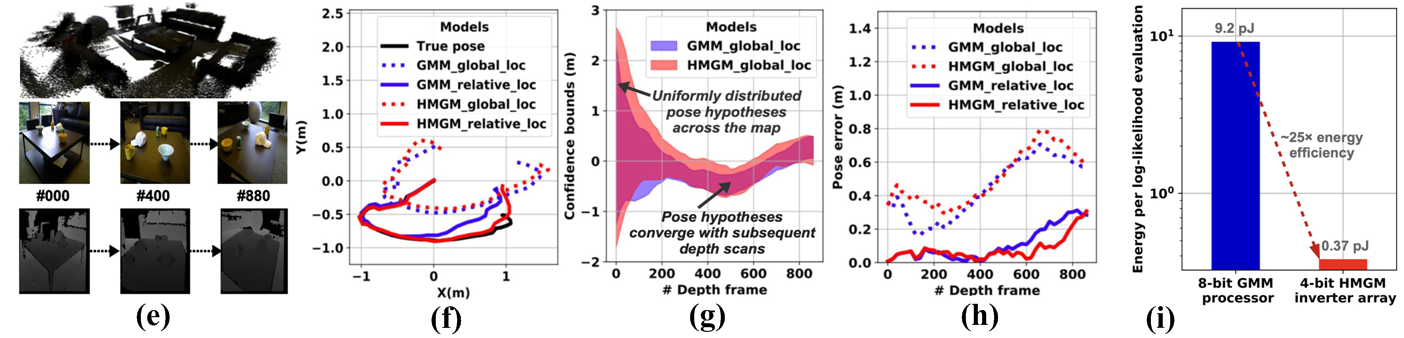}
    \caption{\textbf{Compute-in-memory-based ultra-low-power Monte-Carlo localization framework for insect-scale drones.} \textbf{(a)} CIM architecture employs columns of 6-T inverters to leverage their switching current for likelihood evaluations. \textbf{(b)} Gaussian-like switching current of 6-T inverters. \textbf{(c)} Contour and \textbf{(d)} surface plot of switching current at varying input voltages showing rectilinear tails of the implemented function whereas a Gaussian function has elliptical tails. On RGB-D dataset \textbf{(e)}, \textbf{(f)}-\textbf{(h)} show the comparison of localization steps between the proposed co-designed HMGM-based approach vs. conventional GMM-based approach. \textbf{(i)} Comparison of energy between an 8-bit digital processor and 4-bit HMGM inverter array-based CIM.  [Reproduced from our prior work \cite{shukla2021ultralow} under IEEE's open-source license agreement.]}
    \label{fig:syslevel}
\end{figure*}

\section{Compute-in-Memory Particle Filtering-based Drone Localization}
\subsection{Overview of Particle Filtering-based Localization}
By applying Bayes' rule, the posterior probability of the drone's pose ($\mathbf{x}$) is given as $P(\mathbf{x}|\mathbf{z}) = \frac{P(\mathbf{z}|\mathbf{x})P(\mathbf{x})}{P(\mathbf{z})}
$. Here, $P(\mathbf{z}|\mathbf{x})$ represents the measurement likelihood, and $P(\mathbf{x})$ represents the prior probability of the pose. The Maximum Likelihood Estimate (MLE) of the drone's pose seeks to maximize the likelihood of the current measurement, i.e., $
\mathbf{x}_{\text{MLE}} = \text{argmax}~P(\mathbf{z}|\mathbf{x}) $. On the other hand, the Maximum Aposteriori Estimate (MAP) of the pose aims to maximize the posterior probability of the pose, i.e., $
\mathbf{x}_{\text{MAP}} = \text{argmax}~P(\mathbf{z}|\mathbf{x})P(\mathbf{x})$.

When the drone has complete uncertainty about its pose during global localization, MAP and MLE estimates are equivalent. However, MAP estimation provides a systematic way to incorporate prior beliefs when there is partial pose knowledge. In probabilistic localization, the drone can systematically reduce estimation uncertainties by recursively integrating more measurements using a recursive Bayes rule:
\begin{subequations}
\begin{equation}
\overline{bel}(\mathbf{x_t}) = \sum_{x_{t-1}}P(\mathbf{x_t}|\mathbf{u_t},\mathbf{x_{t-1}})bel(\mathbf{x_{t-1}})
\label{eqnpredict}
\end{equation}
\begin{equation}
bel(\mathbf{x_t}) = \eta P(\mathbf{z_t}|\mathbf{x_t})\overline{bel}(\mathbf{x_t})
\label{eqncorrect}
\end{equation}
\end{subequations}
In Eq. (\ref{eqnpredict}), $\overline{bel}(\mathbf{x_t})$ represents the drone's updated pose belief at iteration $t$, and $bel(\mathbf{x_{t-1}})$ is the pose belief from the previous iteration $t-1$. $\mathbf{u_t}$ denotes the control input applied to the drone, and $P(\mathbf{x_t}|\mathbf{u_t},\mathbf{x_{t-1}})$ is the probabilistic motion model, which leverages the Markov assumption, implying that the belief of $\mathbf{x_t}$ depends only on the previous state's belief and the most recent control input $\mathbf{u_t}$.

Particle Filtering, as described in \cite{djuric2003particle}, offers a practical way to implement Bayesian filtering when dealing with arbitrary likelihood and belief function profiles through a Monte Carlo approach. Instead of relying on analytically defined likelihood and belief functions, this method involves considering a collection of hypotheses, often referred to as particles, that are sampled based on the underlying density functions. Each particle is assigned a weight, which is updated based on measurements using the equations mentioned earlier. As measurements are acquired, particles with low likelihood values are systematically removed. To account for various sources of uncertainty, such as the drone's motion control, a new set of particles is regenerated from the current set, which represents various uncertainties such as uncertainties of motion control.

\subsection{Posterior Density Estimation with an Inverter Array}
A commonly used method for obtaining $P(\mathbf{z}|\mathbf{x})$ involves the utilization of a Gaussian mixture model (GMM) to represent the 3D map \cite{reynolds2009gaussian}. By employing 3D scanning devices like the Microsoft Kinect, point-cloud data of operating domains is captured and fitted with a GMM. To accurately capture the intricacies of the 3D map, it is essential to employ an adequate number of mixture components within the GMM \cite{eckart2018fast}.

In a typical assessment of pose localization, the computation of the likelihood, denoted as $p_{\text{GMM}}(\mathbf{z_{proj}}(\mathbf{x},\mathbf{z}); \mathbf{\Theta})$, involves numerous non-zero depth pixels and a substantial number of mixture components within the GMM model. Consequently, the computational burden associated with calculating $P(\mathbf{z}|\mathbf{x})$ becomes quite substantial. Additionally, considering that drones must continually determine their position during flight, there are strict timing constraints placed on the evaluation of $P(\mathbf{z}|\mathbf{x})$. In the following, we discuss how CIM can significantly minimize these processing overheads by exploiting a novel co-design approach.

Consider the design of a six-transistor inverter controlled by three input voltages: V\textsubscript{X}, V\textsubscript{Y}, and V\textsubscript{Z}, as illustrated in Fig. 2(a), shown within the highlighted box. The transistors within the inverter incorporate a floating gate mechanism for programming their threshold voltage in a non-volatile manner, such as using charge trap transistor mechanisms \cite{gu2019charge}. The programmability of the inverter array is achieved by adjusting the charge density within the floating gate. In Fig. 2(b), the current flowing through the series-connected transistors within the inverter, denoted as I$_\text{INV}$, effectively emulates a Gaussian-like function. The plot illustrates the behavior of I$_\text{INV}$ when one of the input voltages, namely V\textsubscript{X}, V\textsubscript{Y}, or V\textsubscript{Z}, is varied while keeping the other fixed.

Figs. 2(c, d) depicts contour and surface plots of I\textsubscript{INV} as two gate voltages vary simultaneously. The overall current of a multi-input inverter can be approximated as 1/(1/I\textsubscript{INV,1} + 1/I\textsubscript{INV,2} + 1/I\textsubscript{INV,3}), which is equivalent to the harmonic mean of the constituent inverters' currents. When controlling the inverter's current with multiple input voltages, the characteristics of I\textsubscript{INV} are more accurately represented by a Harmonic Mean of Gaussian-like (HMG) function, where each constituent 1D function is influenced by input voltages V\textsubscript{X}, V\textsubscript{Y}, and V\textsubscript{Z}. Notably, compared to implementing and evaluating a multivariate Gaussian function using a digital datapath, the approach presented in Fig. 2(a) is significantly simpler. In a digital design, evaluating a multivariate Gaussian requires digital components such as multipliers, adders, and lookup tables for exponential or log-ADD functions. In contrast, the HMG function is realized with just six transistors and can be evaluated by applying analog voltages V\textsubscript{X}, V\textsubscript{Y}, and V\textsubscript{Z}. 

\subsection{Compute-in-Memory Particle Filtering}
As discussed in our prior work \cite{shukla2021ultralow}
, based on the current pose belief, the scan $\mathbf{z}$ of $N$ non-zero depth map pixels $\{z_1, z_2, ...,z_N\}$ is projected to 3D via the camera's projection model which determines the inputs V\textsubscript{X}, V\textsubscript{Y}, and V\textsubscript{Z}  applied at the inverter array in Fig. 2(a). Evaluating the likelihood of projections thereby determines the likelihood of the pose-belief, i.e., how likely the current belief on the drone's pose is based on the measurements and flying domain's map model. In a typical evaluation of pose localization, the likelihood is computed for hundreds of non-zero depth pixels and hundreds of mixture functions in the GMM model; therefore, has an excessive workload. Meanwhile, using the above methodology where map models are learned using a mixture of HMGs and implemented on a floating gate inverter array, the likelihood of each projection can be readily implemented. The total current from the parallel inverters effectively emulates the likelihood of measurements applied at the gates. This net current is then converted into the digital domain using a logarithmic Analog-to-Digital Converter for post-processing. 

Using the simplified co-design approach, Figs. 2(e-h) compares the prediction accuracy of drone localization with our HMG mixtures (HMGM)-based approach against the conventional GMM-based model on the RGB-D Scenes Dataset v2 \cite{lai2014unsupervised}. Notably, the co-designed approach achieves a matching accuracy to the conventional approach. Although CIM-based implementation necessitates data converters, ADC and DAC, the overhead associated with these components amortizes efficiently when applied in a larger-scale processing architecture consisting of numerous parallel columns. By leveraging these unique characteristics of our CIM hardware and co-design methodologies, in our prior work \cite{shukla2021ultralow} and Fig. 2(i), the total energy consumption (with 500 inverter columns emulating 100 mixture components) for the likelihood estimation was estimated to be 374 fJ (at 45 nm CMOS process technology), which is 25$\times$ lower than an 8-bit GMM processor. 

\section{Compute-in-Memory Uncertainty-Expressive Bayesian Visual Odometry}
\subsection{Overview of Visual Odometry (VO)}
In Visual Odometry (VO), the camera's ego motion is determined by analyzing changes in visual data captured during its operation. Traditional techniques relied on recognizing and tracking distinctive visual features like corners or edges in the camera's field of view. Recent advancements in VO involve the utilization of deep neural networks (DNN) to directly learn the relationship between the camera's visual input and its ego-motion from empirical data. PoseNet \cite{kendall2015posenet} pioneered the training of convolutional neural networks for end-to-end tracking of monocular camera ego-motion. PoseLSTM \cite{walch2017} leveraged long-short-term memories (LSTM) to enhance ego-motion accuracy compared to frame-based approaches such as PoseNet. DeepVO \cite{wang2017} showcased the power of recurrent convolutional neural networks (RCNNs) in capturing features and modeling sequential dependencies between successive camera frames. UnDeepVO \cite{li2018} introduced unsupervised deep learning technique for recovering absolute scale. VLocNet \cite{valada2018,radwan2018} presented a multitask monocular model that leverages parameter sharing and auxiliary learning for VO.

\subsection{Bayesian and Variational Inference of Neural Networks}
While most deep learning-based frameworks for VO result in a point prediction of the vehicle's pose and cannot express their prediction confidence, to improve the predictive robustness, a Bayesian approach to deep learning's inference is gaining attention which can determine both the point estimate of pose as well as confidence on the prediction. Notably, in contrast to classical neural networks, where network weights and predictions are deterministic point estimates, Bayesian neural networks (BNNs) represent network weights as a statistical distribution, ushering in two key advantages over their classical counterparts. \textit{Firstly}, BNNs rest on a more solid theoretical foundation by considering a distribution of weights. Learning from data is inherently an ill-posed problem and may yield an infinite number of possible weight solutions. Classical neural networks rely on a single-point estimate for weights, overlooking other potential weight combinations. In contrast, BNNs explore numerous weight configurations, leading to more robust predictions. \textit{Secondly}, unlike classical neural networks, BNNs can express the confidence of their predictions. BNN predictions take the form of a probabilistic distribution, with higher-order moments like variance providing insight into the prediction confidence.

However, the predictive robustness of BNNs also demands an overwhelming number of computations when compared to a classical neural network. A practical implementation of BNN operates by sampling many weights from the posterior weight distribution. At each sample, the network output is computed and weighed against the sample probability. The cumulative output is obtained only after a sufficiently large number of samples ($\sim$100 – 1000) are extracted and the network output at each sample is computed. Since the number of weights in even a moderate-sized network ranges from hundreds to tens of thousands, both sampling in the high-dimensional weight space and the output extraction of each sample is computationally expensive. Hence, without speeding up the calculations, BNN remains inaccessible for real-time applications.

To minimize the complexity of Bayesian inference, a variational approximation of the posterior can be utilized. In Variational Inference (VI), a family of approximate probability density functions, denoted as $\mathcal{F}$, is considered. Within this family, a specific member $q^*(w)$ is sought that minimizes the Kullback-Leibler (KL) divergence from the true posterior distribution. Subsequently, the posterior density $P(w \, | \, D)$ is approximated using this optimal member $q^*(w)$. Previous studies \cite{blei2017variational} have discussed algorithms for obtaining $q^*(w)$. However, a critical requirement is to select a flexible family $\mathcal{F}$ that allows $q^*(w)$ to closely approximate the true posterior. 

\begin{figure*}[t]
    \centering
    \includegraphics[width=\linewidth]{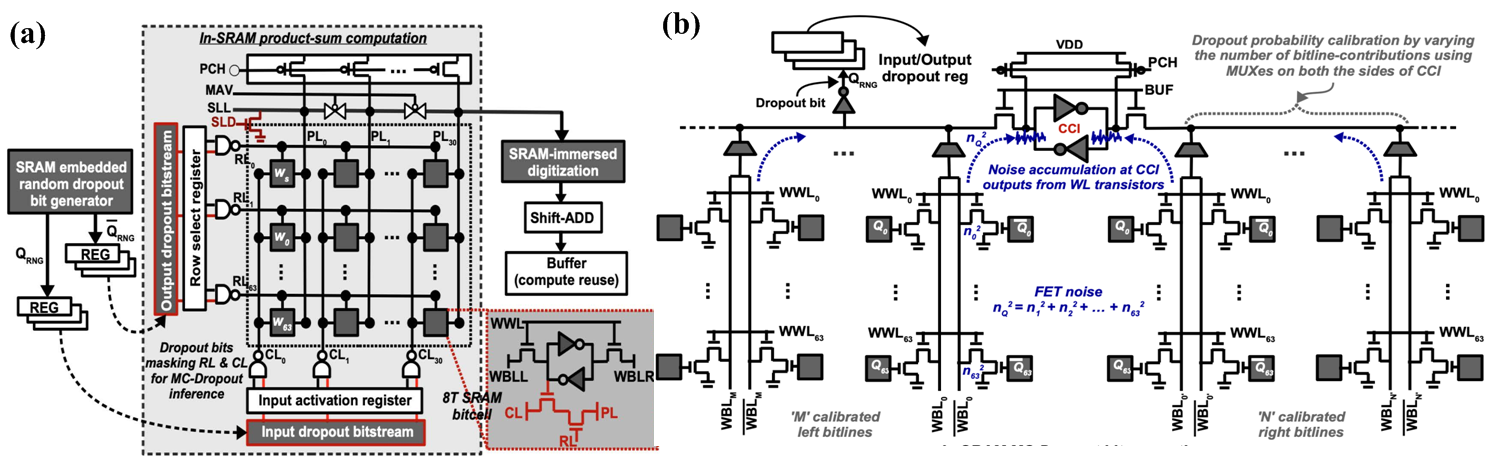}
    \includegraphics[width=\linewidth]{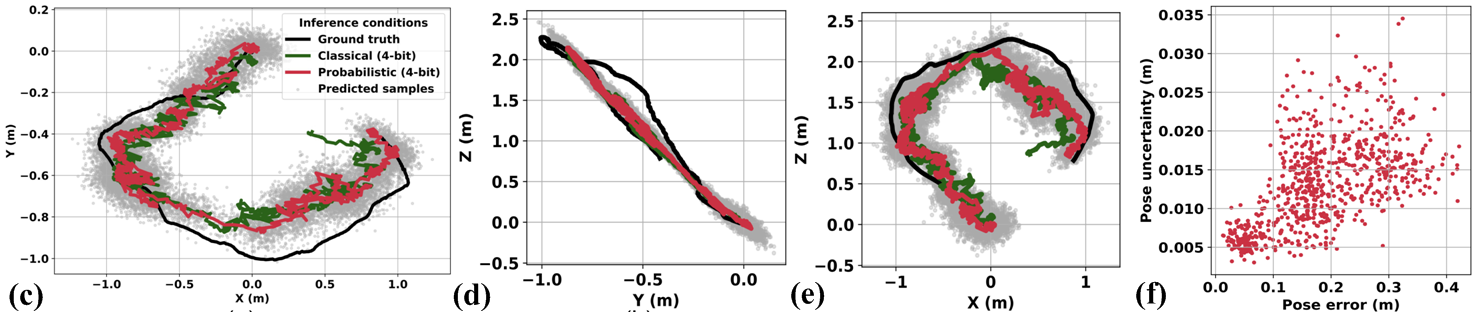}
    \caption{\textbf{Compute-in-memory-based acceleration of variational inference of neural networks:} \textbf{(a)} An SRAM-based CIM macro combines storage and Bayesian inference (BI) functionalities. The inset figure illustrates an 8T SRAM cell featuring both storage and product ports. The CIM incorporates a randomly generated dropout bit generator, enabling MC-Dropout inference. \textbf{(b)} CIM-embedded dropout bit generator. MC-Dropout for indoor (RGB-D dataset) drone localization and trajectory estimation in \textbf{(c)} X-Y, \textbf{(d)} Y-Z, and \textbf{(e)} X-Z position coordinates. We also compare these results with deterministic network configurations under various inference conditions. \textbf{(f)} The correlation between uncertainty (variance) and the error in the pose estimates. [Reproduced from \cite{shukla2022mc} with IEEE's permission]}
    \label{fig:syslevel}
\end{figure*}

\subsection{Compute-in-Memory Monte-Carlo Dropout-based}
Monte Carlo Dropout (MC-Dropout), an adaptation of variational inference, was introduced in \cite{gal2016dropout} and is adopted here for CIM-based acceleration. In a dropout (DO) layer, input and output neurons are randomly excluded in each iteration, following a Bernoulli distribution. Employing MC-dropout layers with a dropout probability of 0.5 has been demonstrated to effectively capture model uncertainties for robust inference across various applications \cite{kendall2016modelling, gal2016dropout}. Additionally, in \cite{gal2017concrete}, authors have devised learning methods to determine optimal dropout probabilities from training data. Predictions in this MC-Dropout-based inference procedure involve averaging model outputs for regression tasks and majority voting for classification tasks. Model confidence is extracted by estimating the variance of each iteration's output.

While CIM has shown tremendous benefits for high-dimensional scalar products of weight matrices and input vectors, for probabilistic inference procedures such as MC-Dropout, additional capabilities are needed. In Fig. 3(a), to facilitate random input dropouts, the inputs to CL peripherals undergo an AND operation with a dropout bitstream. Similarly, for random output dropouts, row activations are masked by ANDing RL signals with an output dropout bitstream. Consequently, in MC-Dropout, an extra step is required to generate dropout bits for each applied input vector, making the high-speed generation of these dropout bit vectors a critical overhead for CIM-based MC-Dropout \cite{shukla2022mc}.

Fig. 3(b) illustrates the proposed SRAM-embedded random number generator (RNG), which takes advantage of SRAM's write parasitics for SRAM-immersed generation of dropout bits. During inference, write wordlines (WWL) to a CIM macro are deactivated. Consequently, along a column, each write port injects leakage and noise current into the bit line. Despite variations in leakage current from each port, denoted as $I_{leak,ij}$, due to threshold voltage ($V_{TH}$) mismatches, a notable phenomenon occurs. The cumulative leakage current across parallel ports reduces the overall sensitivity of the net leakage current at the bit lines. In other words, $\sum_i I_{leak,ij}$ exhibits lower sensitivity to $V_{TH}$ mismatches. Each write port also contributes noise current, $I_{noi,ij}$, to the bit line. Since the noise current from each port varies independently, the net noise current is amplified. 

We exploit this filtering of process-induced mismatches and magnification of noise sources at the bit lines to implement a within-SRAM cross-coupled inverter (CCI)-based random number generator (RNG). In Fig. 3(b), an equal number of SRAM columns are connected to both ends of the CCI. Both bit lines (BL and $\overline{\text{BL}}$) of a column are connected to the same end to cancel out the effect of column data. Both ends of the CCI are precharged using a precharge (PCH) signal and then discharged using column-wise leakage currents for half a clock cycle at the clock transition, pull-down transistors are activated using a delayed PCH to generate the dropout bit. For calibration, the CCI generates a fixed number of output random bits serially, allowing for the estimation of its bias. 

Moreover, since two consecutive iterations of MC-Dropout share some input/output neurons, this can allow for a substantial reduction in computational effort by iteratively calculating the sum of products in each iteration using $P_{i} = P_{i-1} + W\times I^A_{i} - W\times I^D_{i}$. Here, $I^A_{i}$ represents the input neurons that are activated in the current iteration but were inactive in the preceding one, while $I^D_{i}$ signifies the input neurons that were active in the preceding iteration but are inactive in the current one. Employing this strategy enables the system to conduct only the new calculations while efficiently reutilizing the computations that overlap between iterations. In our prior work \cite{shukla2022mc}, we further discussed how the processing flow order can be optimized to even more substantially reduce the necessary processing. 

\subsection{Uncertainty Expressive Visual Odometry (VO)}
Figs. 3(c-e) demonstrates the performance of the above framework for VO while utilizing the RGB-D Scenes Dataset v2 \cite{lai2014unsupervised}. The comparison between ground truth and estimated pose trajectories highlights that even with very low precision, probabilistic inference can accurately track the ground truth. Fig. 3(f) shows a scatter plot correlating pose error with variance (uncertainty) in probabilistic inference where there is a discernible correlation between error and predictive uncertainty. This indicates that, in contrast to classical (deterministic) inference, CIM's probabilistic inference can signal potential mispredictions by exhibiting high predictive uncertainties. In our prior work \cite{shukla2022mc}, we also characterized the energy efficiency of the above framework while operating at a clock frequency of 1 GHz and a voltage of 0.85V, and designing the system at the 16 nm CMOS node. The developed framework was benchmarked to operate at 3.04 TOPS/W (at 4-bit precision) and around 2 TOPS/W (at 6 bits) for 30 MC-Dropout network iterations. Such high efficiency throughout of Monte Carlo-based Bayesian framework demonstrates the potential of predictive uncertainty-aware robust inference at the edge. 

\section{Conclusions and Future Works}
We have presented ultra-low-power CIM-based prediction frameworks that effectively express uncertainty, such as CIM-based particle filtering for drone localization. Additionally, we explored CIM-based processing of variational inference in deep learning models, harnessing intrinsic shot noise for robust probabilistic processing. Central to both frameworks is our co-design methodologies that adapts the traditional computing models to CIM's operating and physical constraints for significant simplicity and parallelism in processing while leveraging analog mode representations to minimize the workload to begin with. However, a notable challenge in these uncertainty-expressive prediction frameworks is their reliance on Monte Carlo methods, which can be complex and resource-intensive. To address this, future research could investigate Monte Carlo-free uncertainty-aware frameworks like conformal inference \cite{gibbs2021adaptive,stutts2023lightweight, stutts2023mutual} and evidential learning \cite{sensoy2018evidential} suited for the edge. 

\bibliographystyle{IEEEtran}
\bibliography{main}
\end{document}